%% file: jmlr.tex
\documentclass[twoside,11pt]{article}

\usepackage{blindtext}

\usepackage{jmlr2e}

\usepackage{float}
\usepackage{sidecap}
\usepackage{multirow}

\usepackage{url}            
\usepackage{booktabs}       
\usepackage{amsfonts}       
\usepackage{nicefrac}       
\usepackage{microtype}      
\usepackage{graphicx}

\usepackage{lastpage}
\jmlrheading{26}{2025}{1-\pageref{LastPage}}{3/24; Revised
12/24}{1/25}{24-0439}{Renhao Wang, Yu Sun, Arnuv Tandon, Yossi Gandelsman,
Xinlei Chen,
Alexei A. Efros,
Xiaolong Wang}

\ShortHeadings{Test-Time Training on Video Streams}{Wang, Sun, Tandon, Gandelsman,
Chen,
Efros,
Wang}
\firstpageno{1}
\input{notations}

\begin{document}

\title{Test-Time Training on Video Streams}

\author{\small
Renhao Wang$^*$,
Yu Sun\thanks{Equal contribution. Correspondence to: \url{yusun@berkeley.edu}.\\
Renhao Wang, Yu Sun, Yossi Gandelsman, Alexei A. Efros are with UC Berkeley.
Arnuv Tandon is with Stanford University.
Xinlei Chen is with
Meta AI.
Xiaolong Wang is with
UC San Diego.\\
Project website with videos, dataset and code: \url{https://test-time-training.github.io/video}}~,
\,Arnuv Tandon, Yossi Gandelsman, \\
Xinlei Chen,
Alexei A. Efros,
Xiaolong Wang
}

\editor{Samy Bengio}

\maketitle

\begin{abstract}
Prior work has established Test-Time Training (TTT) as a general framework to further improve a trained model at test time.
Before making a prediction on each test instance, the model is first trained on the same instance using a self-supervised task such as reconstruction.
We extend TTT to the streaming setting, where multiple test instances -- video frames in our case -- arrive in temporal order.
Our extension is online TTT:
The current model is initialized from the previous model, then trained on the current frame and a small window of frames immediately before.
Online TTT significantly outperforms the fixed-model baseline for four tasks, on three real-world datasets.
The improvements are more than 2.2$\times$ and 1.5$\times$ for instance and panoptic segmentation.
Surprisingly, online TTT also outperforms its offline variant that accesses strictly more information, training on all frames from the entire test video regardless of temporal order.
This finding challenges those in prior work using synthetic videos.
We formalize a notion of \textit{locality} as the advantage of online over offline TTT, and analyze its role with ablations and a theory based on bias-variance trade-off.
\end{abstract}


\begin{figure}[h!]
    \centering
    \vspace{1ex}
    \includegraphics[width=0.95\textwidth]{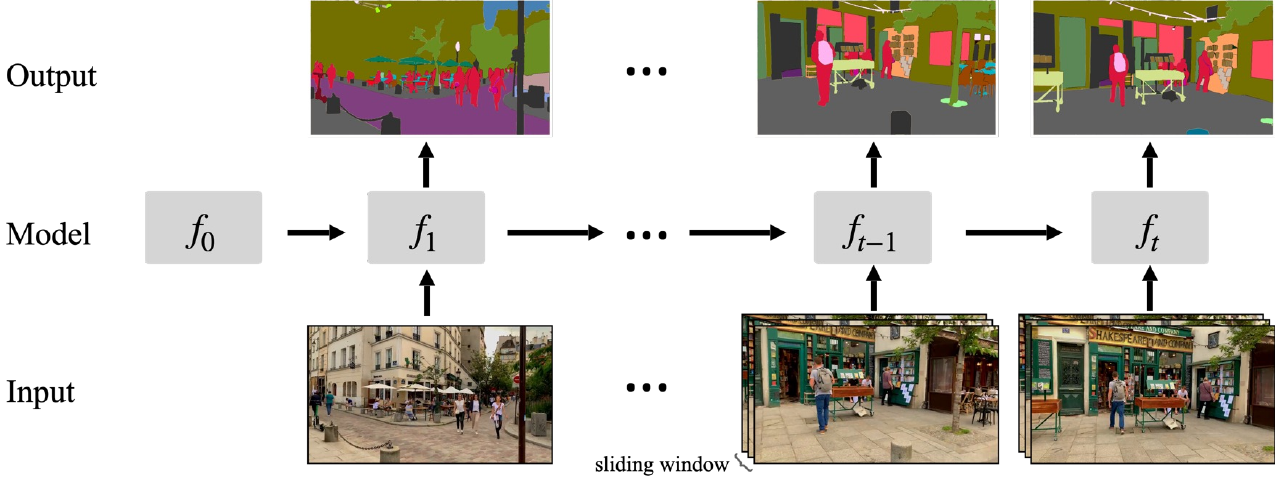}
    \caption{\small
    In our streaming setting, the current model $f_t$ makes a prediction on the current frame before it can see the next one. The prediction task here is segmentation.
    $f_t$ is obtained through online TTT, initializing from the previous model $f_{t-1}$. 
    Each video is treated as an independent unit.
    A sliding window of size $k$ contains the current and previous frames as test-time training data for the self-supervised task.
    Concretely, $k = 16$ gives a window of only 1.6 seconds in our experiments.
    }
    \label{fig-method}
\end{figure}

\clearpage

\begin{figure}
    \centering
    \begin{minipage}{0.53\textwidth}
        \includegraphics[width=\textwidth]{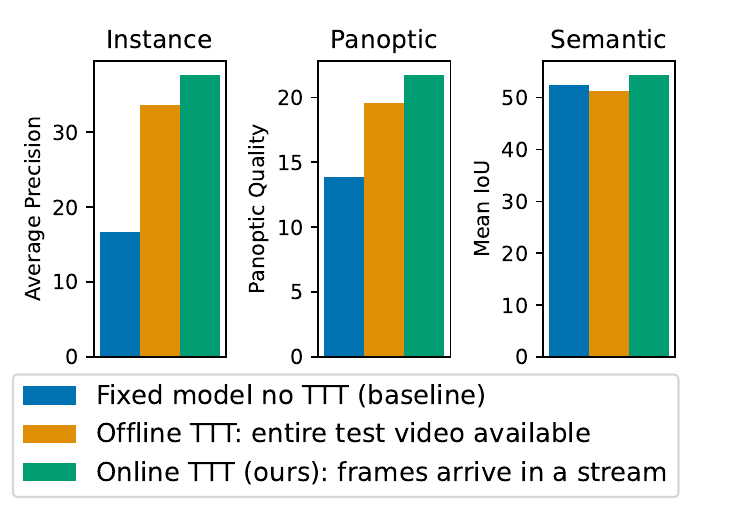}
    \end{minipage}
    \begin{minipage}{0.46\textwidth}
        \caption{\small
            Results for instance and panoptic segmentation on COCO Videos, and semantic segmentation on KITTI-STEP.
            Online TTT (green) performs the best, and only requires the realistic setting where the video frames arrive in a stream.
            Offline TTT (yellow) requires the rather unrealistic setting where all frames from the entire test video are available before making predictions.
            We think of offline TTT as ``training on all possible futures''.
            Still, online TTT outperforms offline by taking advantage of locality.
            Details in Section~\ref{analysis}.
        }
        \label{fig-bars}
    \end{minipage}
\end{figure}

\section{Introduction}
\label{intro}
Most models in machine learning today are fixed during deployment.
As a consequence, a trained model must prepare to be robust to all possible futures. 
This can be hard because being ready for all futures limits the model’s capacity to be good at any particular one, even though only one future actually happens.
The basic idea of Test-Time Training~(TTT) is to continue training on the future once it arrives in the form of a test instance \citep{sun2020test}. Since each test instance is observed without a ground truth label, training is performed with self-supervision.

This paper investigates TTT on video streams, where each ``future'', or test instance $x_t$, is a frame, and each video is treated as an independent unit.
Naturally, $x_t$ and $x_{t+1}$ are visually similar.
We focus on the intuition discussed above: 
Is training on the future once it actually happens better than training on all possible futures?
More concretely, this question in the streaming setting asks about the role of \emph{locality}:
For making a prediction on a particular frame $x_t$, is it better to perform test-time training \emph{offline} on all of $x_1,\dots,x_T$ (the video ends at time $T$),
or \emph{online} on only $x_t$ (and maybe a few previous frames)?

Our empirical evidence supports locality.
The best performance is achieved through online TTT: For each $x_t$, only train on itself, and a small window of less than two seconds of frames immediately before $t$.
We call this sliding window of frames the \emph{explicit memory}.
The optimal explicit memory needs to be short term -- in plain language, some amount of \textbf{forgetting is actually beneficial}.
This high-level finding challenges those from prior work in TTT~\citep{wang2020tent, volpi2022road} 
and continual learning
\citep{li2017learning, lopez2017gradient, kirkpatrick2017overcoming},
but is consistent with recent work in neuroscience \citep{gravitz2019importance}.

For online TTT, parameters after training on $x_t$ carry over as initialization for training on $x_{t+1}$.
We call this the \emph{implicit memory}.
Because such an initialization is usually quite good to begin with, most of the benefit from online TTT is realized with only one gradient step per frame.
Not surprisingly, the effectiveness of both explicit and implicit memory depends on \emph{temporal smoothness} -- that $x_t$ and $x_{t+1}$ are similar.
In Section \ref{analysis}, we conduct ablations on both explicit and implicit memory, and develop a theory based on bias-variance trade-off under smoothness.

Experiments in this paper are also of practical interests, besides conceptual ones.
Models for many computer vision tasks are trained with large datasets of still images,
e.g. COCO, for segmentation, but deployed on video streams. 
The default is to naively run such models frame-by-frame,
since temporal smoothing (i.e. averaging across a sliding window of predictions) offers little improvement.
Online TTT significantly improves prediction quality on three real-world video datasets, for four tasks: 
semantic, instance and panoptic segmentation, and colorization.
Figure~\ref{fig-bars} visualizes results for the first three tasks (since metrics for colorization are less reliable):
\textbf{online TTT beats even the offline oracle.}

We also collect a new video dataset with dense annotations -- COCO Videos. These videos are orders of magnitude longer than in other public datasets, and contain much harder scenes from diverse daily-life scenarios.
Longer and more challenging videos better showcase the importance of locality, making it even harder to perform well on all futures at once.
The improvements on COCO Videos are, respectively, more than 2.2$\times$ and 1.5$\times$ for instance and panoptic segmentation.

One of the most popular forms of self-supervision in computer vision is reconstruction: removing parts of the input image, then predicting the removed content \citep{denoisingautoencoder, pathak2016context, beit, simmim}.
Recently, a class of deep learning models called masked autoencoders (MAE) \citep{mae}, using reconstruction as the self-supervised task, has been highly influential.
TTT-MAE \citep{ttt-mae} adopts these models for test-time training using reconstruction. 
The main task in \cite{ttt-mae} is object recognition.
Inspired by the empirical success of TTT-MAE, we use it as a subroutine inside online TTT, 
and extend it to other main tasks such as segmentation.

Prior work \citep{sun2020test} experiments with online TTT (without explicit memory) in the streaming setting, but each $x_t$ is drawn independently from the same test distribution.
This test distribution is created by adding some synthetic corruption, e.g. Gaussian noise, 
to a test set of still images, e.g. ImageNet test set \citep{imagenet_c}.
Therefore, all $x_t$s belong to the same ``future'', and locality is meaningless:
TTT on as many $x_t$s as possible achieves the best performance by learning to ignore the corruption.
TTT on actual video streams is fundamentally different and much more natural.

More recently, \cite{volpi2022road} also experiments in the streaming setting.
While the $x_t$s here are not independent, these videos are short clips, again simulated to contain synthetic corruptions, e.g. CityScapes with Artificial Weather.
Therefore, like in \cite{sun2020test}, each corruption moves all $x_t$s into almost the same ``future'', which they call a domain. 
Since performance drop is caused by that shared corruption, it is best recovered by training on all $x_t$s.
Their only dataset without corruptions (CityScapes) sees little improvement 
(1.4\% relative to no TTT).
There is no mentioning of locality, our basic concept of interest.

\section{Related Work}
\label{related}
\subsection{Continual Learning}
\label{related-cts}

In the field of continual a.k.a. lifelong learning, a model learns a sequence of tasks in temporal order, and is asked to perform well on all of them \citep{van2019three, hadsell2020embracing}.
Here is the conventional setting:
Each task is defined by a data distribution $P_t$, 
which produces a training set $D^{\text{tr}}_t$
and a test set $D^{\text{te}}_t$.
At each time $t$, the model is evaluated on all the test sets $D^{\text{te}}_1, \dots, D^{\text{te}}_t$ of the past and present, and average performance is reported.

The basic solution is to simply train the model on all of $D^{\text{tr}}_1, \dots, D^{\text{tr}}_t$, which collectively have the same distribution as all the test sets.
This is often referred to as the oracle with infinite memory (a.k.a. replay buffer) that remembers everything.
However, due to memory constraints, the model at time $t$ is only allowed to train on $D^{\text{tr}}_t$.
More advanced solutions, therefore, focus on how to retain memory of past data only with model parameters \citep{santoro2016meta, li2017learning, lopez2017gradient, shin2017continual, kirkpatrick2017overcoming, gidaris2018dynamic}.

Some of the literature extends beyond the conventional setting.
\cite{aljundi2019task} uses continuous instead of discrete tasks across time.
\cite{purushwalkam2022challenges} and \cite{fini2022self}
perform self-supervised learning on unlabeled training sets, 
and evaluate the learned features on the test sets.
\cite{hoffman2014continuous}, \cite{li2020online} and \cite{panagiotakopoulos2022online} use 
a labeled training set $D^{\text{tr}}_0$
in addition to unlabeled training sets $D^{\text{tr}}_1, \dots, D^{\text{tr}}_t$, 
connecting with unsupervised domain adaptation.
\cite{diaz2018don} uses alternative metrics, e.g. forward transfer, to justify forgetting for reasons other than computational.

Much of continual learning is motivated by the hope to understand human memory and generalization through the lens of artificial intelligence \citep{hassabis2017neuroscience, de2021continual}.
Our work shares the same motivation, but focuses on test-time training, without distinct splits of training and test sets.

\subsection{Test-Time Training}
\label{related-ttt}
One of the earliest algorithms for training at test time is \cite{bottou1992local}:
For each test input, train on its neighbors before making a prediction.
This approach continues to be effective for support vector machines (SVM)~\citep{zhang2006svm} and recently in large language models~\citep{hardt2023test}.
\cite{bottou1992local}, titled \emph{Local Learning}, articulates locality as a basic concept in machine learning.
Another line of work called transductive learning uses test data to add constraints to the margin of SVMs~\citep{joachims2002learning, collobert2006large, vapnik2013nature}.
The principle of transduction, as stated by Vapnik, also emphasizes locality ~\citep{Gammerman98learningby, vapnik_book}:
``Try to get the answer that you really need but not a more general one."

In computer vision, the idea of training at test time has been well explored for specific applications~\citep{jain2011online, shocher2018zero, nitzan2022mystyle, xie2023sepico}, 
especially depth estimation~\citep{tonioni2019learning, tonioni2019real, zhang2020online, zhong2018open, luo2020consistent}.
Our paper extends TTT-MAE \citep{ttt-mae}, detailed in Section \ref{background}.
TTT-MAE, in turn, is inspired \cite{sun2020test},
which proposed the general framework for test-time training with self-supervision.
The particular self-supervised task used in \cite{sun2020test} is rotation prediction \citep{gidaris2018unsupervised}.
Many other papers have followed this framework since then
\citep{hansen2020self, sun2021online, liu2021ttt++, yuan2023robust}, including \cite{volpi2022road} on videos discussed in Section \ref{intro}, and \cite{azimi2022self} which we discuss next.

In \cite{azimi2022self}, each video is treated as a dataset of unordered frames instead of a stream.
In particular, there is no concept of past vs. future frames.
The same model is used on the entire video.
In contrast, our paper emphasizes locality.
We have access to only the current and past frames, and our model keeps learning over time.
In addition, all of our results are on real world videos, while \cite{azimi2022self} experiment on videos with artificial corruptions.
These corruptions are also i.i.d. across frames.

Our paper is very much inspired by~\cite{mullapudi2018online}. 
To make video segmentation more efficient, their paper makes predictions frame-by-frame using a small student model.
If the student is not confident, it queries an expensive teacher model, and then trains the student to fit the prediction from the teacher online.
Thanks to temporal smoothness, the student can generalize confidently across many frames without querying the teacher,
so learning and predicting combined is still faster than naively using the teacher at every frame.
Our method only consists of one model, which learns from a self-supervised task instead of a teacher model.
Rather than focusing on computational efficiency as in~\cite{mullapudi2018online}, the main goal of our paper is to improve inference quality.
Behind their particular algorithm, however, we see the shared idea of locality, regardless of the form of supervision.

\begin{figure*}
  \begin{center}
    \includegraphics[width=\textwidth]{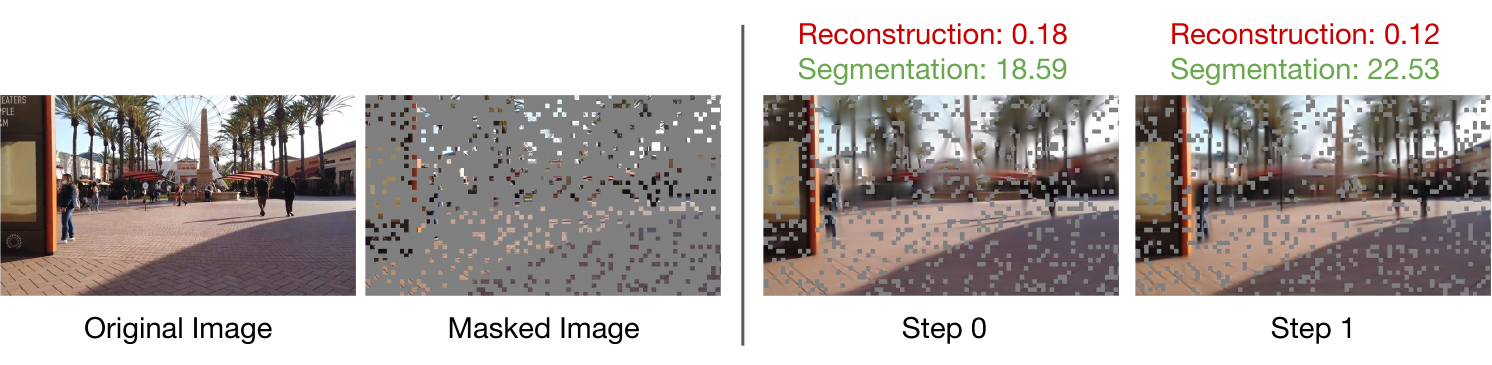}
  \end{center}
  \vspace{-1ex}
  \caption{\small
  Training a masked autoencoder (MAE) to reconstruct each test image at test time. Reconstructed images on the right visualize the progress of gradient descent on this one-sample learning problem. For each test image, TTT-MAE first masks out majority of the patches. The masked image is given as input to the autoencoder, which then reconstructs those masked patches. The reconstruction loss is the pixel-wise mean squared error between the original and reconstructed patches. Loss on the main task -- panoptic segmentation -- also falls as reconstruction gets better.
  The unmasked patches are not shown on the right since they are not part of the reconstruction loss.
}
  \label{fig-mae}
\end{figure*}

\section{Background: TTT-MAE}
\label{background}
Our paper extends the work of \emph{Test-Time Training with Masked Autoencoders} (TTT-MAE) \citep{ttt-mae}, and uses TTT-MAE as the subroutine that updates the model for each frame.
This section briefly describes TTT-MAE, as background for our extension.
Figure \ref{fig-mae} illustrates the process of TTT-MAE.

The general architecture for TTT with self-supervision \citep{sun2020test} is Y-shaped with a stem and two heads: 
a prediction head $g$ for the self-supervised task,
another prediction head $h$ for the main task, 
and a feature extractor $f$ as the stem.
The output features of $f$ are shared between $g$ and $h$ as input.
For TTT-MAE, the self-supervised task is masked image reconstruction \citep{mae}.
Following standard terminology for autoencoders, $f$ is also called the encoder, and $g$ the decoder.

Each input image $x$ is first split into many non-overlapping patches. 
To produce the autoencoder input $\tilde{x}$, we mask out majority, e.g. 80\%, of the patches in $x$ at random.
The self-supervised objective $\ell_s(g \circ f(\tilde{x}), x)$ compares the reconstructed patches from $g \circ f(\tilde{x})$ to the masked patches in $x$, and computes the pixel-wise mean squared error.
For the main task, e.g. segmentation, all patches in the original $x$ are given as input to $h \circ f$, during both training and testing.

\subsection{Training-Time Training}
There are three widely accepted ways to optimize the model components ($f$, $g$, $h$) at training time: joint training, probing, and fine-tuning.
Fine-tuning is unsuitable for TTT, because it makes $h$ rely too much on features that are used by the main task. 
Our paper uses joint training, described in Section \ref{method}.
In contrast, \cite{ttt-mae} uses probing, which we describe next for completeness.

To prepare for probing, the common practice is to first train $f$ and $g$ with $\ell_s$ on the training set without ground truth.
This preparation stage is also called self-supervised pre-training.
\cite{ttt-mae} uses the encoder and decoder already pre-trained by \cite{mae}, denoted by $f_0$ and $g_0$.
During probing, the main task head $h$ is then trained separately by optimizing for $\ell_m(h \circ f_0(x), y)$, on the training set with ground truth.
$f_0$ is kept frozen.
We denote $h_0$ as the main task head after probing.
Since $h_0$ has been trained for the main task using features from $f_0$ as input, $h_0 \circ f_0$ can be directly applied on each test image as a baseline without TTT, keeping the parameters of $f_0$ and $h_0$ fixed.

\subsection{Test-Time Training}
At test time, TTT-MAE takes gradient steps on the following one-sample learning problem:
\begin{equation}
\label{optimize_test}
f', g' = \argmin_{f, g}
l_s(g \circ f( \tilde{x}'), x'),
\end{equation}
then makes the final prediction $h_0 \circ f'(x')$.
Crucially, the gradient-based optimization process always starts from $f_0$ and $g_0$.
When evaluating on a test set, \cite{ttt-mae} always discards $f'$ and $g'$ after making a prediction on each test input $x'$, and resets the weights back to $f_0$ and $g_0$ for the next test input. 
By test-time training on the test inputs independently, \cite{ttt-mae} does not assume that they can help each other.

In the original MAE design \citep{mae}, $g$ is very small relative to $f$, and only the visible patches, e.g. 20\%, are processed by $f$.
Therefore the overall computational cost of training for the self-supervised task in only a fraction, e.g. $25\%$, of training for the main task.
In addition to speeding up training-time training for reconstruction, this reduces the extra test-time cost of TTT-MAE.
Each gradient step at test time, counting both forward and backward, costs only half the time of forward prediction for the main task.

\section{Test-Time Training on Video Streams}
\label{method}

We consider each test video as a smoothly changing sequence of frames $x_1,\ldots,x_T$; time $T$ is when the video ends.
Each video is treated as an independent unit.
In the streaming setting, an algorithm is evaluated on the video following its temporal order, like how a human would consume it.
At each time $t$, the algorithm should make a prediction on $x_t$ after receiving it from the environment, before seeing any future frame.
In addition to $x_t$, the past frames $x_1,\ldots,x_{t-1}$ are also available at time $t$, if the algorithm chooses to use them.
Ground truth labels are never given to the algorithm on test videos.

Now we describe our algorithm for this streaming setting.
At a high level, our algorithm simply amounts to a loop over the video frames, wrapped around TTT-MAE \citep{ttt-mae}.
In practice, making it work involves many design choices.

\subsection{Training-Time Training}
At training time, if there was no self-supervision, then it is straightforward to optimize $h \circ f$ end-to-end for the main task only. The trained model produced by this process can already be applied on each $x_t$ without TTT. In Table \ref{tab-main}, we call this baseline \emph{Main Task Only}.
But such a model is not suitable for TTT, since then the self-supervised head $g$ would have to be trained from scratch at test time.

To make $g$ well-initialized for TTT, at training time we jointly optimize all three model components in a single stage, end-to-end, on both the self-supervised task and main task. This is called joint training.
While joint training was also an option for prior work on TTT-MAE~\citep{ttt-mae}, empirical experience at the time indicated that probing performed better (see Section \ref{background}).
In this paper, however, we have successfully tuned joint training to be as effective as probing, and therefore default to joint training because it is simpler than the two-stage process of probing. 

Following the notations in Section~\ref{background}, the self-supervised task loss is denoted by $\ell_s$, and the main task loss is $\ell_m$.
During joint training, we optimize those two losses together to produce a self-supervised task head $g_0$, main task head $h_0$, and feature extractor $f_0$:
\begin{align*}
g_0, h_0, f_0 =
\argmin_{g, h, f}
\frac{1}{n}\sum_{i=1}^{n} 
[
&\ell_m(h \circ f(x_i), y_i)\\
&+ \ell_s(g \circ f(\tilde{x}_i), x_i)
].
\end{align*}
The summation is over the training set with $n$ samples, each consisting of input $x_i$ and label $y_i$.
As discussed in Section~\ref{background},
$\tilde{x}_i$ is $x_i$ transformed as input for the self-supervised task. In the case of MAE, $\tilde{x}_i$ is obtained by masking $80\%$ of the patches in $x_i$.
Note that although the test instances come from video streams, training-time training uses labeled, still images, 
e.g. in the COCO training set, 
instead of unlabeled videos.

After joint training, the fixed model $h_0 \circ f_0$ can also be applied directly on each $x_t$ without TTT, just like for \emph{Main Task Only}. 
We call this new baseline \emph{MAE Joint Training}.
Empirically, these two baselines have roughly the same performance. Joint training does not hurt or help when only considering the fixed model after training-time training.

\subsection{Test-Time Training}
\label{ttt-method}
Another baseline is to blithely apply TTT-MAE by plugging each test frame $x_t$ as $x'$ into Equation \ref{optimize_test}, following the process in Section~\ref{background}.
We call this ablation \emph{TTT-MAE No Memory} in Table~\ref{tab-main} and Table~\ref{tab-ablation}.
In this ablation,
TTT for every $x_t$ is initialized with $h_0$ and $f_0$, by resetting the model parameters back to those after joint training.
Like \emph{Main Task Only} and \emph{MAE Joint Training}, this ablation misses the point of using a video.
All three baselines treat each video as a collection of unordered, independent frames that might not contain any information about each other.
None of the three can improve over time, no matter how long a video explores the same environment.

Improvement over time is only possible through some form of memory, by retaining information from the past frames $x_1,\ldots,x_{t-1}$ to help prediction on $x_t$.
Because evaluation is performed at each timestep only on the current frame, our memory design should favor past data that are most relevant to the present.
Fortunately, with the help of nature, the most recent frames usually happen to be the most relevant due to \emph{temporal smoothness} -- observations that are close in time tend to be similar.
We design memory that favors recent frames in the following two ways.

\textbf{Implicit memory.}
The most natural improvement is to simply not reset the model parameters between timesteps.
That is, to initialize test-time training at timestep $t$ with $f_{t-1}$ and $g_{t-1}$, instead of $f_0$ and $g_0$.
This creates an implicit memory, since information carries over from the previous parameters, which was already optimized on previous frames.
It also happens to be more biologically plausible: we humans do not constantly reset our minds.
In prior work \citep{sun2020test},
TTT with implicit memory is called the ``online'' version, in contrast to the ``standard'' version with reset, for the baseline setting of independent images without temporal smoothness discussed in the paragraphs above.

\textbf{Explicit memory.}
A more explicit way of remembering recent frames is to keep them in a sliding window.
Let $k$ denote the window size. At each timestep $t$, our method solves the following optimization problem instead of Equation \ref{optimize_test}:
\begin{equation}
\label{ttt-loss}
f_t, g_t = \argmin_{f, g} \frac{1}{k}~~\sum_{t'=t-k+1}^{t} 
 \ell_s(g \circ f(\tilde{x}_{t'}), x_{t'}),
\end{equation}
before predicting $h_0 \circ f_t(x_t)$.
Optimization is performed with stochastic gradient descent:
at each iteration, we sample a batch with replacement, uniformly from the same window.
Masking is applied independently within and across batches.
It turns out that only one iteration is sufficient for our final algorithm, because given temporal smoothness, implicit memory should already provide a good initialization for the optimization problem above.



\subsection{Implementation Details}
In principle, our method is applicable to any architecture.
Our current implementation uses Mask2Former \citep{cheng2021masked}, which has achieved state-of-the-art performance on many semantic, instance and panoptic segmentation benchmarks.
Our Mask2Former uses a Swin-S \citep{liu2021swin} backbone -- in our case, this is also the shared encoder $f$. Everything following the backbone in the original architecture is taken as the main task head $h$, and our decoder $g$ copies the architecture of $h$ except the last layer that maps into pixel space for reconstruction.
Joint training starts from their model checkpoint, 
which has already been trained for the main task. 
Only $g$ is initialized from scratch.

Following \cite{mae}, we split each input into patches, and mask out 80\% of them. 
However, unlike the Vision Transformers \citep{dosovitskiy2020image} used in \cite{mae}, 
Swin Transformers use convolutions.
Therefore, we must take the entire image as input (with the masked patches in black) instead of only the unmasked patches.
Following \cite{pathak2016context}, we use a fourth channel of binaries to indicate if the corresponding input pixels are masked.
The model parameters for the fourth channel are initialized from scratch before joint training.
If a completely transformer-based architecture for segmentation becomes available in the future, our method would like become even faster, by not encoding the masked patches \citep{mae, ttt-mae}.

\renewcommand{\arraystretch}{1.1}
\begin{table}
\centering
\scalebox{0.88}{
\begin{tabular}{|l|l|c|c|c|c|c|}
\toprule
\multirow{2}{*}{\textbf{Setting}} &
\multirow{2}{*}{\textbf{Method}} &
\multicolumn{2}{c|}{\textbf{COCO Videos}} &
\multicolumn{3}{c|}{\textbf{KITTI-STEP}}\\
& & \emph{Instance} & \emph{Panoptic}
& \emph{Val.} & \emph{Test} & \emph{Time}\\
\midrule
\multirow{3}{*}{Independent frames} & Main Task Only & 16.7 & 13.9 & 53.8 & 52.5 & 1.8 \\
& MAE Joint Training & 16.5 & 13.5 & 53.5 & 52.5 & 1.8 \\
& TTT-MAE No Memory & 35.4 & 20.1 & 53.6 & 52.5 & 3.8 \\
\midrule
Entire video available 
& Offline TTT-MAE All Frames & 33.6 & 19.6 & 53.2 & 51.2 & 1.8 \\
\midrule
\multirow{6}{*}{Frames in a stream} &
LN Adapt                    & 16.5 & 14.7 & 53.8 & 52.5 & 2.0 \\  
& Tent                      & 16.6 & 14.6 & 53.8 & 52.2 & 2.8 \\  
& Tent with Class Balance                & 16.7 & 14.8 & 53.8 & 52.5 & 3.7 \\
& Self-Train                               & - & - & 54.7 & 54.0 & 6.6 \\ 
& Self-Train with Class Balance          & - & - & 54.1 & 53.6 & 6.9 \\
\cline{2-7}
\rule{0pt}{3ex}
& Online TTT-MAE (Ours) & \textbf{37.6} & \textbf{21.7} & \textbf{55.4} & \textbf{54.3} & 4.1 \\
\bottomrule
\end{tabular}
}
\caption{\small
Metrics for instance, panoptic and semantic segmentation are, respectively, average precision (AP), panoptic quality (PQ), and mean IoU (\%). 
Time is in seconds per frame, using a single A100 GPU, averaged over the KITTI-STEP test set. Time costs on COCO Videos are similar, thus omitted for clarity.
The self-training baselines are not applicable for instance and panoptic segmentation because the model does not return a confidence per object instance.
Bars in Figure~\ref{fig-bars} correspond to values from the following rows: blue for \emph{Main Task Only}, yellow for \emph{Offline TTT-MAE All Frames}, and green for \emph{Online TTT-MAE (Ours)}.
}
\label{tab-main}
\end{table}

\section{Results}
We experiment with four applications on three real-world datasets:
1) semantic segmentation on KITTI-STEP -- a public dataset of urban driving videos;
2) instance and panoptic segmentation on COCO Videos -- a new dataset we annotated;
3) colorization on COCO Videos and a collection of black and white films.
Please visit our project website at \url{https://video-ttt.github.io/} to watch videos of our results.

\subsection{Additional Baselines}
In Section \ref{method}, we have discussed three baselines:
\emph{Main Task Only}, \emph{MAE Joint Training}, and \emph{TTT-MAE No Memory}. 
We now discuss other baselines. 
Some of these baselines actually contain our own improvements.

\paragraph{Alternative architectures.}
The authors of Mask2Former did not evaluated it on KITTI-STEP.
We benchmark Mask2Former on the KITTI-STEP validation set against two other popular models of comparable size:
SegFormer B4~\citep{xie2021segformer} (64.1M), and 
DeepLabV3+/RN101~\citep{chen2017rethinking} (62.7M),
which is used by \cite{volpi2022road}.
Their mean IoUs are, respectively, 42.0\% and 53.1\%. 
Given that \emph{Main Task Only} in Table~\ref{tab-main} has 53.8\%, we can verify that our pre-trained model (69M) is indeed the state-of-the-art on KITTI-STEP.
For COCO segmentation, the authors of Mask2Former have already compared with alternative architectures~\citep{cheng2021masked}, so we do not repeat their experiments.

\paragraph{Majority vote with augmentation.}
We also experiment with test-time augmentation of the input,
applying the default data augmentation recipe in the codebase for 100 predictions per frame, then taking the majority vote across predictions as the final output.
This improves \emph{Main Task Only} by 1.2\% mean IoU on the KITTI-STEP validation set.
Combining the same technique with our method yields roughly the same improvement, indicating that they are again orthogonal.
For clarity, we do not use majority vote elsewhere in this paper.

\paragraph{Temporal smoothing.}
We implement temporal smoothing by averaging the predictions across a sliding window, in the same fashion as our explicit memory. 
The window size is selected to optimize performance after smoothing on the KITTI-STEP validation set.
This improves \emph{Main Task Only} by only 0.4\% mean IoU.
Applying temporal smoothing to our method also yields 0.3\% improvement.
This indicates that our method is orthogonal to temporal smoothing.
For clarity, we do not use temporal smoothing elsewhere in this paper.

\paragraph{Alternative techniques for TTT.}
Self-supervision with MAE is only one particular technique for test-time training.
Subsection \ref{ttt-method} describes a general loop, and any technique that does not use ground truth labels can be used as a subroutine to update the model inside the loop. 
We experiment with three promising techniques according to prior work: self-training \citep{volpi2022road}, layer norm (LN) adaptation \citep{schneider2020improving}, and Tent \citep{wang2020tent}. 
For self-training, our implementation significantly improves on the version in \cite{volpi2022road}.
Please refer to Appendix \ref{app-baselines} for an in depth discussion of these three techniques.

\paragraph{Class balancing.}
\cite{volpi2022road} proposes a heuristic that can be used in conjunction with implicit memory: Record the number of predicted classes, for the initial model $h\circ f_0$ and the current model $h\circ f_t$. 
Reset the model parameters when the difference is large enough, in which case the predictions of the current model have likely collapsed.
To compare with~\cite{volpi2022road}, we evaluate this heuristic on self-training and Tent.
This heuristic cannot be applied to LN Adapt, which does not actually modify the trainable parameters in the model.

\begin{table}
\centering
\begin{tabular}{|c|cccc|}
\toprule
 Dataset & Length & Frames & Rate & Classes\\\midrule
{\small CityScapes-VPS \citep{kim2020video}}
& 1.8 & 3000 & 17 & 19
\\  \midrule
{\small DAVIS \citep{Pont-Tuset_arXiv_2017}}
& 3.5 & 3455 & 30 & - \\  \midrule
{\small YouTube-VOS \citep{xu2018youtube}}
& 4.5 & 123,467 & 30 & 94 \\  \midrule
{\small KITTI-STEP \citep{weber2021step}}
& 40 & 8,008 & 10 & 19
\\  \midrule
{\small COCO Videos (Ours)}
& 350 & 10,475 & 10 & 134 
\\
\bottomrule
\end{tabular}
\caption{\small
Video datasets with annotations for segmentation. 
The columns are: average length per video in seconds, total number of frames in the entire dataset, rate in frames per second, and total number of classes.
COCO Videos is larger than KITTI-STEP in total frames. The 3 videos in our new dataset are roughly an order of magnitude longer than those in KITTI-STEP, and much more diverse in terms of number of classes. In fact, each of the 3 videos alone contains more frames than all of the videos combined in the KITTI-STEP validation set.
}
\label{tab-datasets}
\end{table}

\subsection{Semantic Segmentation on KITTI-STEP}
\label{emp-kitti}
KITTI-STEP \citep{weber2021step} contains 9 validation videos and 12 test videos of urban driving scenes.\footnote{
KITTI-STEP was originally designed to benchmark instance-level tracking, and has a separate test set held-out by the organizers. 
The official website evaluates only tracking-related metrics on this test set.
Therefore, we perform our own evaluation using the segmentation labels.
Since we do not perform regular training on KITTI-STEP, we use the training set as test set.
}
At the rate of 10 frames-per-second, these videos are the longest -- up to 106 seconds -- among public datasets with dense pixel-wise annotations.
All hyper-parameters, even for COCO Videos, are selected on the KITTI-STEP validation set.
Joint training is performed on CityScapes~\citep{cordts2016cityscapes}, another driving dataset with exactly the same 19 categories as KITTI-STEP, but containing still images instead of videos.

Table \ref{tab-main} presents our main results.
Figure \ref{fig-kitti} in the appendix visualizes predictions on two frames.
Please see project website for more visualizations.
\emph{Online TTT-MAE} in the streaming setting, using both implicit and explicit memory, performs the best.
For semantic segmentation, such an improvement is usually considered highly significant in the community.

Baseline techniques that adapt the normalization layers alone do not help at all in these evaluations.
This finding agrees with the evidence in \cite{volpi2022road}: LN Adapt and Tent help significantly on datasets with synthetic corruptions, but do not help on real-world dataset (e.g. CityScapes).

\emph{Online TTT-MAE} optimizes for only one iteration per frame, and turns out slower than the baselines without TTT by $2.3\times$. 
Comparing with \cite{ttt-mae}, which optimizes for 20 iterations per frame (image), our method runs much faster, again, because implicit memory takes advantage of temporal smoothness to get a better initialization for every frame.
Resetting parameters is wasteful on videos, because the adjacent frames are very similar.

\subsection{COCO Videos}
\label{emp-coco}

While KITTI-STEP already contains the longest annotated videos among publicly available datasets at the time of submission, they are still far too short for studying long-term phenomenon in locality.
KITTI-STEP videos are also limited to driving scenarios, a small subset of the diverse scenarios in our daily lives.
These limitations motivate us to collect and annotate our own dataset of videos.

We collected 3 videos, each about 5 minutes, annotated by professionals, in the same format as for COCO instance and panoptic segmentation \citep{lin2014microsoft}.
The benchmark metrics are also the same as in COCO: 
average precision (AP) for instance and panoptic quality (PQ) for panoptic.
To put things into perspective, each of the 3 videos alone contains more frames, at the same rate, than all of the videos combined in the KITTI-STEP validation set.
We compare this new dataset with other publicly available ones in Table~\ref{tab-datasets}.


All videos are egocentric, similar to the visual experience of a human walking around. 
In particular, they do not follow any tracked object like in Oxford Long-Term Tracking \citep{valmadre2018long} or ImageNet-Vid \citep{shankar2021image}.
Objects leave and enter the camera view all the time.
Unlike KITTI-STEP and CityScapes that focus on self-driving scenes, 
our videos are both indoor and outdoor.

We start with the publicly available Mask2Former model pre-trained on still images in the COCO training set. Analogous to our procedure for KITTI-STEP, joint training for TTT-MAE is also on COCO images,
and our 3 videos are only used for evaluation.
Mask2Former is the state-of-the-art on the COCO validation set, with 44.9~AP for instance and 53.6~PQ for panoptic segmentation. But its performance in Table \ref{tab-main} drops to 16.7~AP and 13.9~PQ on COCO Videos. This highlights the challenging nature of COCO Videos, and the fragility of models trained on still images when evaluated on videos in the wild.

We use exactly the same hyper-parameters as tuned on the KITTI-STEP validation set, for all algorithms considered.
That is, all of our results for COCO Videos were completed in a single run.
As it turns out in Figure~\ref{fig-window}, using a larger window size would further improve performance. 
However, we believe such hyper-parameters for TTT should not be tuned on the test videos, so we stick to the window size selected on the KITTI-STEP validation set.

Table \ref{tab-main} presents our main results.
Comparing to \emph{Main Task Only}, our relative improvements for instance and panoptic segmentation are, respectively, more than 2.2$\times$ and 1.5$\times$.
Improvements of this magnitude on the state-of-the-art is usually considered dramatic.
The self-training baselines are not applicable here because for instance and panoptic segmentation, the model does not return a confidence per object instance.

Interestingly, \emph{TTT-MAE No Memory} also produces notable improvements on both tasks, and even outperforms \emph{Offline TTT-MAE All Frames}.
Considering this result from the perspective of locality, \emph{Offline TTT-MAE All Frames} is the most global method, since it tries to be good at all frames in each video. 
At the other end of the spectrum, \emph{TTT-MAE No Memory} is the most local, since it only uses information from the current frame.
For COCO Videos, local is better than global, if one has to pick an extreme.

\begin{table*}
\centering
\scalebox{0.88}{
\begin{tabular}{|l|l|l|l|l|l|}
\toprule
\textbf{Method} & \textbf{FID}~$\downarrow$~~~ & \textbf{IS}~$\uparrow$ & 
\textbf{LPIPS}~$\uparrow$ & \textbf{PSNR}~$\uparrow$ & \textbf{SSIM}~$\uparrow$ 
\\
\midrule
Zhang et al.~\citep{zhang2016colorful} & 62.39 & 5.00 $\pm$ 0.19 & 0.180 & 22.27 & \textbf{0.924}  \\
Main Task Only \citep{cheng2021masked} & 59.96 & 5.23 $\pm$ 0.12 &  0.216 & 20.42 & 0.881  \\
\midrule
Online TTT-MAE (Ours) & \textbf{56.47} & \textbf{5.31 $\pm$ 0.18} & \textbf{0.237} & \textbf{22.97} & 0.901 \\
\bottomrule
\end{tabular}
}
\caption{\small
Quantitative results for video colorization on COCO Videos. FID: Fr\'echet Inception Distance. IS: Inception Score (standard deviation is naturally available).
LPIPS: Learned Perceptual Image Patch Similarity. PSNR: Peak Signal-to-Noise Ratio. SSIM: Structural Similarity.
Arrows pointing up indicate higher the better, and pointing down indicate lower the better.
}
\label{tab-colorization}
\end{table*}

\subsection{Video Colorization}
\label{emp-color}
The goal of colorization is to add realistic RGB colors to gray-scale images ~\citep{lei2019fully, zhang2019deep}.
Our goal here is to demonstrate the generality of our method, not to achieve the state-of-the-art.

Following \cite{zhang2016colorful}, we simply treat colorization as a supervised learning problem. 
We use the same architecture as for segmentation -- Swin Transformer with two heads, pre-trained on ImageNet -- to predict colors given gray-scale images.
We do not use domain-specific techniques, e.g., perceptual losses, adversarial learning, or diffusion.
Our bare-minimal baseline already achieves results comparable to those in \cite{zhang2016colorful}.
\emph{Online TTT-MAE} uses exactly the same hyper-parameters as for segmentation.
All of our colorization experiments were completed in a single run.
Because colorizing COCO Videos is expensive, we only evaluate \emph{Online TTT-MAE} and \emph{Main Task Only}.

For the quantitative results in Table \ref{tab-colorization}, we colorize COCO Videos by first processing the 3 videos into black and white. This enables us to compare with the original videos in RGB.
For qualitative results, we also colorize the 10 original black-and-white Lumiere Brothers films from 1895, roughly 40 seconds each, at the rate of 10 frames per second.
Figure~\ref{fig:colorization} in Appendix \ref{app-color} provides a snapshot of our qualitative results.
Please also see Appendix \ref{app-color} for a list of the films and their lengths.

Our method outperforms the baseline and \cite{zhang2016colorful} on all metrics except SSIM.
It is a field consensus that PSNR and SSIM often misrepresent actual visual quality because colorization is inherently multi-modal \citep{zhang2016colorful,zhang2017real}, but we still include them for completeness.
Please see the project website for the complete set of the original and colorized videos.
Our method visually improves the quality in all of them comparing to the baseline, especially in terms of consistency across frames.

\begin{table}
\centering
\scalebox{0.9}{
\begin{tabular}{|l|c|c|c|c|}
\toprule
\multirow{2}{*}{\textbf{Method}} &
\multicolumn{2}{c|}{\textbf{COCO Videos}} &
\multicolumn{2}{c|}{\textbf{KITTI-STEP}}\\
& \emph{Instance} & \emph{Panoptic}
& \emph{~~~~Val.~~~~} & \emph{~~~~Test~~~~}\\
\midrule
TTT-MAE No Memory    & 35.4 & 20.1 & 53.6 & 52.5 \\
\midrule
Implicit Memory Only                & 36.1 & 20.7 & 54.3 & 54.4 \\
Explicit Memory Only                & 35.7 & 20.2 & 53.6 & 52.5 \\
\midrule
Online TTT-MAE (Ours)             & {37.6} & {21.7} & {55.4} & {54.3} \\
\bottomrule
\end{tabular}
}
\caption{\small
Ablations on our two forms of memory.
For ease of reference, the values in Table \ref{tab-main} for \emph{TTT-MAE No Memory} and \emph{Online TTT-MAE (Ours)} are reproduced here.
}
\label{tab-ablation}
\end{table}

\section{Analysis on Locality}
\label{analysis}

Now we come back to the two philosophies presented at the beginning of our introduction:
training on all possible futures in advance vs. training on the future once it actually happens.
In other words, training globally vs. locally.


\subsection{Empirical Analysis}
\label{forget-ablate}
Table \ref{tab-ablation} contains ablations on our two forms of memory: implicit and explicit. Both forms of memory contribute to the improvement of \emph{Online TTT-MAE} over \emph{TTT-MAE No Memory} \citep{ttt-mae}. Beyond these basic ablations, we further ablate three aspects of our method.

\paragraph{Offline TTT-MAE.}
This ablation, presented at the beginning of the paper as the yellow bars in Figure \ref{fig-bars}, trains a single model for each test video. 
It lives in a new setting, where all frames from the entire test video are available for training with the self-supervised task, e.g. MAE, before predictions are made on that video.
Strictly more information is provided here than the streaming setting, where only current and past frames are available.
The frames are shuffled into a training set, and gradient iterations are taken on batches sampled from this training set, in the same way as sampled from the sliding window in \emph{Online TTT-MAE}.
To give \emph{Offline TTT-MAE All Frames} even more advantage, we report results from the best iteration on each test video, as measured by {actual test performance}, which would not be available in real world. 
For many videos, this best iteration number is around 1000.

\begin{figure*}
  \vspace{-2ex}
  \begin{center}
    \includegraphics[width=\textwidth]{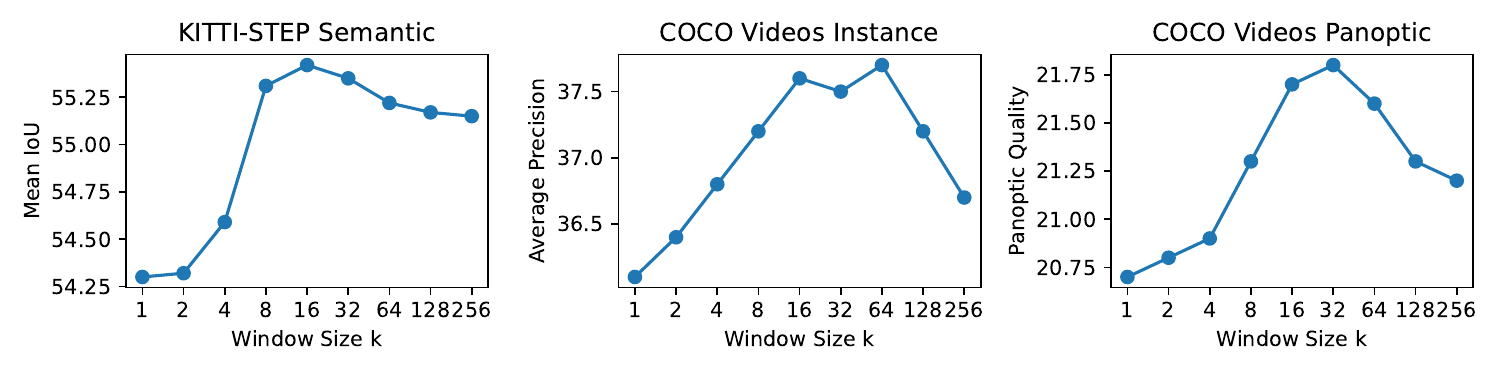}
  \end{center}
  \vspace{-3ex}
  \caption{\small
  Effect of window size $k$ on performance. In simple terms, \emph{Online TTT-MAE} prefers a very short-term memory.
  For all window sizes, the batch size, and therefore computational cost, is fixed. 
  The plot for KITTI-STEP is on the validation set, where we selected the optimal hyper-parameter $k=16$.
  For all three tasks, with a rate of 10 frames per second, 16 frames cover only 1.6 seconds.
  The optimal $k$ on COCO Videos turns out to be different for both semantic and panoptic segmentation, but the results we report in Table~\ref{tab-main} still use $k=16$. The y-values for window size $k=1$ match those for \emph{Implicit Memory Only} in Table~\ref{tab-ablation}.
  Note that the x-axis here is in log scale, in order to highlight the effect of $k$ over a large range. Performance is actually not sensitive to $k$ in linear scale.
}
  \label{fig-window}
\end{figure*}

\paragraph{Window size.}
The choice of whether to use explicit memory is not binary.
On one end of the spectrum, window size $k=1$ is the same as \emph{Implicit Memory Only}.
On the other end, $k=\infty$ comes close to \emph{Offline TTT-MAE All Frames}, except that the future frames cannot be trained on for $k=\infty$.
Figure \ref{fig-window} analyzes the effect of window size on performance.
We observe that too little memory hurts, but so does too much.
This observation makes intuitive sense: frames further in the past become less relevant on average for making a prediction on the current frame, even though they provide more data for TTT.
Figure \ref{fig-locality} illustrates this intuition.

\paragraph{Temporal smoothness.}
As discussed in Section \ref{method}, temporal smoothness is the key assumption that makes our two forms of memory effective.
While this assumption is intuitively necessary, we can verify its importance by shuffling all the frames within each video, therefore destroying temporal smoothness, and observing how results change.
By construction, all three methods under the setting of independent frames -- \emph{Main Task Only}, \emph{MAE Joint Training}, and \emph{TTT-MAE No Memory} --  are not affected.
The same goes with \emph{Offline TTT-MAE All Frames}, which already shuffles the frames during offline training.
For \emph{Online TTT-MAE}, however, shuffling hurts performance dramatically. Performance on the KITTI-STEP validation set becomes worse than \emph{Main Task Only}.

\subsection{Theoretical Analysis}
\label{theory}
To complement our empirical observation that locality can be beneficial, we now rigorously analyze the effect of our window size $k$ for TTT using any self-supervised task.

\paragraph{Notations.}
We first define the following functions of the shared model parameters $\theta$:
\begin{align}
    \nabla \ell_m^{~t}(\theta) &:= \nabla_\theta \ell_m(x_t, y_t; \theta),\\
    \nabla \ell_s^{~t}(\theta) &:= \nabla_\theta \ell_s(x_t; \theta).
\end{align}
These notations have appeared in Section \ref{background} and \ref{method}, where the main task loss $\ell_m$ is defined for object recognition or segmentation, and the
self-supervised task loss $\ell_s$ is instantiated as pixel-wise mean squared error for image reconstruction; $\theta$ refers to parameters of the encoder $f$.

\paragraph{Problem statement.}
Taking gradient steps with $\nabla \ell_m^{~t}$ directly optimizes the test loss, since $y_t$ is the ground truth label of test input $x_t$.
However, $y_t$ is not available, so TTT optimizes the self-supervised loss $\ell_s$ instead.
Among the available gradients, $\nabla \ell_s^{~t}$ is the most relevant.
But we also have the past inputs $x_1, \dots, x_{t-1}$. Should we use some, or even all of them?

\begin{figure*}
  \begin{center}
    \includegraphics[width=\textwidth]{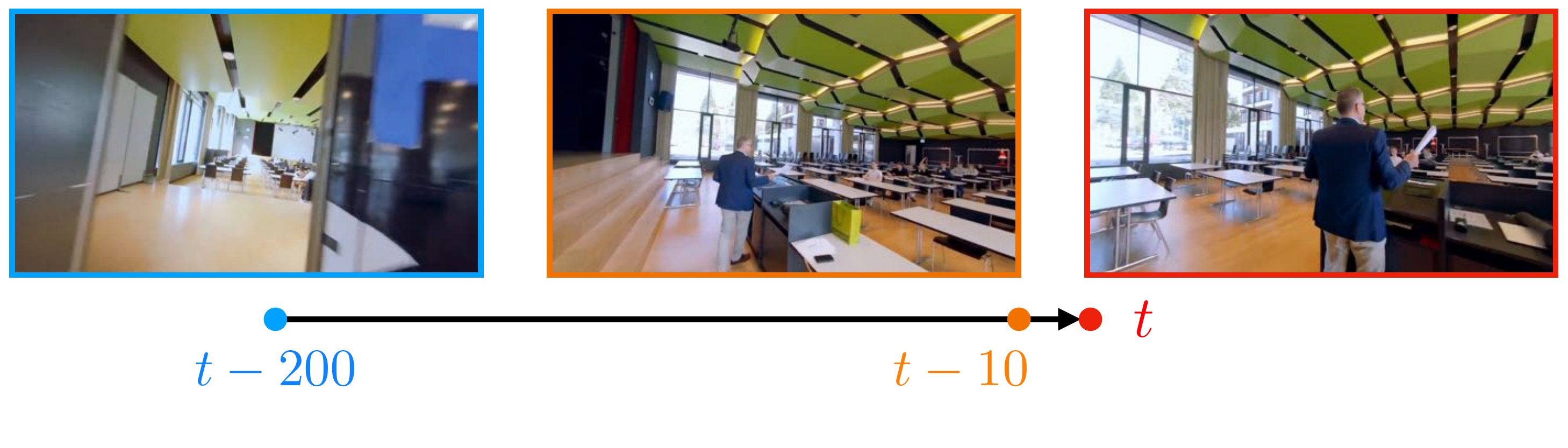}
  \end{center}
  \vspace{-4ex}
  \caption{\small
  An illustration of the principle of locality in video streams. Our goal is improve prediction on the current frame, shot inside a lecture hall. The frame at $t-10$ was still inside this hall. Including this frame in our sliding window decreases variance for TTT. However, the frame at $t-200$ was shot before entering the hall. Including it would significantly increase bias, because it is no longer relevant to the current frame.
  }
  \label{fig-locality}
\end{figure*}

\paragraph{Theorem.}
For every timestep $t$, consider TTT with gradient-based optimization using:
\begin{equation}
\frac{1}{k}
\sum_{t'=t-k+1}^t
\nabla\ell_s^{~t'},
\end{equation}
where $k$ is the window size.
Let $\theta_0$ denote the initial condition, and $\tilde{\theta}$ where optimization converges for TTT.
Let $\theta^*$ denote the optimal solution of $\ell_m^{~t}$ in the local neighborhood of $\theta_0$.
Then we have
\begin{equation*}
\E\left[
\ell_m(x_t, y_t; \tilde{\theta}) - \ell_m(x_t, y_t; \theta^*)
\right]
\leq 
\frac{1}{2\alpha}
\left(
k^2\beta^2\eta^2 + \frac{1}{k}\sigma^2
\right),
\end{equation*}
under the following three assumptions: 
\begin{enumerate}
\item In a local neighborhood of $\theta^*$,
$\ell_m^{~t}$ is $\alpha$-strongly convex in $\theta$, and $\beta$-smooth in $x$.
\item $\|x_{t+1} - x_t\| \leq \eta$.
\item 
$\nabla \ell_m^{~t} = \nabla \ell_s^{~t} + \delta_t$, where $\delta_t$ is a random variable with mean zero and variance $\sigma^2$.
\end{enumerate}
The proof is in Appendix \ref{app-proof}.

\paragraph{Remark on assumptions.}
Assumption 1, that neural networks are strongly convex around their local minima, is widely accepted in the learning theory community \citep{pmlr-v97-allen-zhu19a, zhong2017recovery, wang2021hidden}.
Assumption 2 is simply temporal smoothness in $L^2$ norm; any other norm could also be used here as long as the norm in Assumption 1 for strong convexity is changed accordingly.
Assumption 3, that the main task and self-supervised task have correlated gradients, comes from the theoretical analysis of \cite{sun2020test}.

\paragraph{Bias-variance trade-off.}
Disregarding the constant factor of $1/\alpha$, the upper bound in Theorem 1 is the sum of two terms: $k^2\beta^2\eta^2$ and $\sigma^2/k$.
The former is the bias term, growing with $\eta$. 
The latter is the variance term, growing with $\sigma^2$.
More memory, i.e., sliding window with larger size $k$, reduces variance, but increases bias. This is consistent with our intuition in Figure \ref{fig-locality}.
Optimizing this upper bound w.r.t. $k$ reveals the theoretical sweet spot
$$k = \left( \frac{\sigma^2}{\beta^2\eta^2} \right)^{1/3}.$$

\section{Discussion}
In the end, we connect our work to other ideas in machine learning.

\paragraph{Unsupervised domain adaptation.}
The setting of \emph{Offline TTT-MAE All Frames}, where the entire unlabeled test video is available at once, is very similar to unsupervised domain adaptation (UDA).
Each test video can be viewed as a target domain, and
offline MAE practically treats the frames as i.i.d. data drawn from a single distribution.
The only difference with UDA is that the unlabeled video serves as both training and test data.
In fact, the modified version of UDA above is sometimes called \emph{test-time adaptation}.
Our results suggest that this perspective of seeing each video as a target domain might be misleading for algorithm design, because it discourages locality.

\paragraph{Continual learning.}
Conventional wisdom in the continual learning community believes that forgetting is harmful. 
Specifically, the best accuracy is achieved by remembering everything with an infinite replay buffer, given unlimited computer memory.
Our streaming setting is different from those commonly studied by the continual learning community, because it does not have distinct splits of training and test sets, as explained in Subsection~\ref{related-cts}.
However, our sliding window can be viewed as a replay buffer, and limiting its size can be viewed as a form of forgetting.
In this context, our results suggest that forgetting can actually be beneficial.


\paragraph{TTT on nearest neighbors.}
Here is an alternative heuristic for TTT:
For each test instance, retrieve its nearest neighbors from a training set, and fine-tune the model on those neighbors
before making a prediction on the test instance.
This simple but effective heuristic has been explored in \cite{bottou1992local} and \cite{hardt2023test}, as discussed in Subsection~\ref{related-ttt}.
Given temporal smoothness --  that proximity in time translates to proximity in the retrieval metric, our sliding window can be seen as retrieving ``neighbors" of the current frame.
The only difference is that our ``neighbors" are from the unlabeled test video instead of a labeled trainings set. This difference has two consequences. On one hand, we have to use self-supervision. On the other hand, our ``neighbors" are still relevant (given temporal smoothness) even when the test instance is not represented by the training set.

\paragraph{In-context learning.}
In theory, each test video can be used as the context of a Transformer or RNN, both of which often exhibit the ability of ``in-context learning''~\citep{brown2020language}. As long as the model is autoregressive, the video is still processed as a stream. 
But in practice, this approach requires the model to be already trained on videos.
Our approach, on the other hand, does not use videos as training data, as our very goal is to study the generalization from still images to videos.

\begin{figure}[t]
\vspace{-1ex}
    \centering
    \begin{minipage}{\textwidth}
    \centering
    \includegraphics[width=\textwidth]{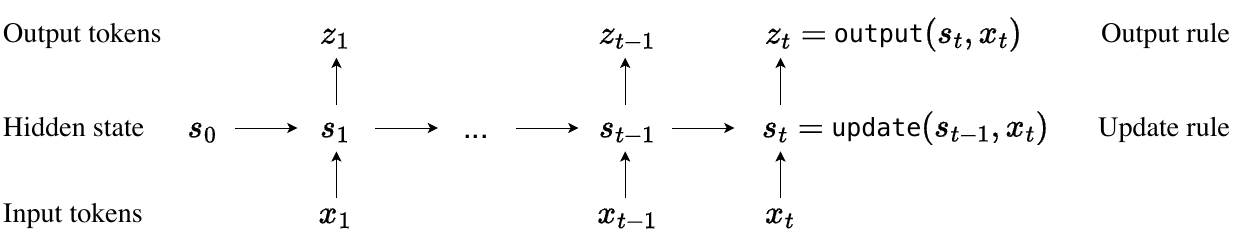}
    \vspace{0.1ex}
    \end{minipage}
    \scalebox{0.865}{
    \renewcommand{\arraystretch}{1.5}
    \begin{tabular}{|l|l|l|l|l|}
    \hline
    & \textbf{Initial state}
    & \textbf{Update rule}
    & \textbf{Output rule}
    & \textbf{Cost}\\
    \hline
    \textbf{Naive RNN} & $s_0 = \texttt{vector()}$ 
    & $s_t = \sigma\left(\theta_{ss}s_{t-1} + \theta_{sx}x_t \right)$ 
    & $z_t = \theta_{zs}s_t + \theta_{zx}x_t$ 
    & $O(1)$ \\
    \textbf{Self-attention} & $s_0 = \texttt{list()}$ 
    & $s_t = s_{t-1}.\texttt{append}(k_t, v_t)$ 
    & $z_t = V_t\texttt{softmax}\left(K^T_tq_t\right)$
    & $O(t)$ \\
    \textbf{Naive TTT} & $W_0 = f.\texttt{params()}$
    & $W_t = W_{t-1} - \eta\nabla\ell(W_{t-1}; x_t)$
    & $z_t = f(x_t; W_t)$
    & $O(1)$ \\
    \hline
    \end{tabular}
    }
    \caption{\small Figure from~\cite{sun2024learning}. \textbf{Top}: A generic sequence modeling layer expressed as a hidden state that transitions according to an update rule.
    All sequence modeling layers can be viewed as different instantiations of three components in this figure: the initial state, update rule and output rule.
    \textbf{Bottom}: Examples of sequence modeling layers and their instantiations of the three components.
    RNN layers compress the growing context into a hidden state of fixed size, therefore their cost per token stays constant.
    Self-attention has a hidden state growing with context, therefore growing cost per token.
    The TTT layer, introduced in \cite{sun2024learning}, is a particular RNN layer with gradient descent as the update rule, following the approach in this paper.
    }
    \label{fig:all-layers}
\end{figure}

\paragraph{Sequence modeling.}
Our method as shown in Figure~\ref{fig-method} closely resembles an RNN as shown in Figure~\ref{fig:all-layers}, if we think of $W$, the parameters of $f$, as the hidden state. From this perspective, online TTT can be regarded as compressing frames $x_1, \dots, x_t$ into $W_t$, and gradient descent is simply a particular update rule.
Following earlier versions of this paper, \cite{sun2023learning} and \cite{sun2024learning} program TTT into sequence modeling layers as an alternative to self-attention, and apply it to language modeling.

\clearpage
\section*{Acknowledgements}
This project is supported in part by Oracle Cloud credits and related resources provided by the Oracle for Research program. Xiaolong Wang’s lab is supported, in part, by NSF CAREER Award IIS-2240014, Amazon Research Award, Adobe Data Science Research Award, and gifts from Qualcomm.
We would like to thank Xueyang Yu and Yinghao Zhang for contributing to the published codebase.
Yu Sun would like to thank his other PhD advisor, Moritz Hardt.

\bibliography{reference}

\appendix

\clearpage
\section{Baseline Techniques for TTT}
\label{app-baselines}


\subsection{Self-Training}
Self-training is a popular technique in  
semi-supervised learning
\citep{Radosavovic_2018_CVPR, rosenberg2005semi, zoph2020rethinking, asano2019self, asano2020labelling}
and domain adaptation \citep{kumar2020understanding, zou2018unsupervised, mei2020instance, liu2021energy, spadotto2021unsupervised}.
It was also evaluated in \cite{volpi2022road} but produced inferior performance.
We experiment with self-training both in its original form, and incorporating our own design improvements.

We assume that for each test image $x$, the prediction $\hat{y}$ is also of the same shape in 2D.
This assumption is satisfied in semantic segmentation and colorization.
We also assume that $F$ outputs a estimated confidence map $\hat{c}$ of the same shape as $\hat{y}$.
Specifically, for pixel $x[i, j]$, $\hat{y}[i, j]$ is the predicted class of this pixel, and $\hat{c}[i, j]$ is the estimated confidence of $\hat{y}[i, j]$.

Self-training repeats many iterations of the following:
\begin{itemize}
\setlength\itemsep{0em}
\setlength{\parskip}{0.5pt}
    \item Start with an empty set of labels $D$ for this iteration.
    \item Loop over every $[i, j]$ location, add pseudo-label $\hat{y}[i, j]$ to $D$ 
    if $\hat{c}[i, j] > \lambda$, 
    for a fixed threshold $\lambda$.
    \item Train $F$ to fit this iteration's set $D$, as if the selected pseudo-labels are ground truth labels.
\end{itemize}
Our first design improvement is the confidence threshold $\lambda$. 
In \cite{volpi2022road}, all predictions are used as pseudo-labels, regardless of confidence.
Our experiments show that for low $\lambda$ or $\lambda=0$ as in \cite{volpi2022road}, self-training is noisy and unstable, as expected.

However, for high $\lambda$, there is limited learning signal, e.g. little gradient, since $f$ is already very confident about the pseudo-label. 
Our second design improvement, inspired by \cite{sohn2020simple}, is to make learning more challenging with an already confident prediction, by masking image patches in $x$.
In \cite{sohn2020simple}, masking is applied sparingly on 2.5\% of the pixels in average. 
We mask 80\% of the pixels, inspired by \cite{mae}.

\subsection{Layer Norm Adapt}
Prior work \citep{schneider2020improving} shows that simply recalculating the batch normalization (BN) statistics works well for unsupervised domain adaptation.
\cite{volpi2022road} applies this technique to video streams by accumulating the statistics with a forward pass on each frame once it is revealed.
Since modern transformers use layer normalization (LN) instead, we apply the same technique to LN.

\subsection{Tent}
The normalization layers (BN and LN) also contain trainable parameters that modify the statistics.
Optimizing those parameters requires a self-supervised objective.
Tent \citep{wang2020tent} is an objective for learning only those parameters at test time, by minimizing the softmax entropy of the predicted distribution over classes.
We update the LN statistics and parameters with Tent, in the same loop as our method, also using implicit and explicit memory.
Hyper-parameters are searched on the KITTI-STEP validation set to be optimal for Tent.

\clearpage
\begin{figure*}
    \centering
    \includegraphics[width=\textwidth]{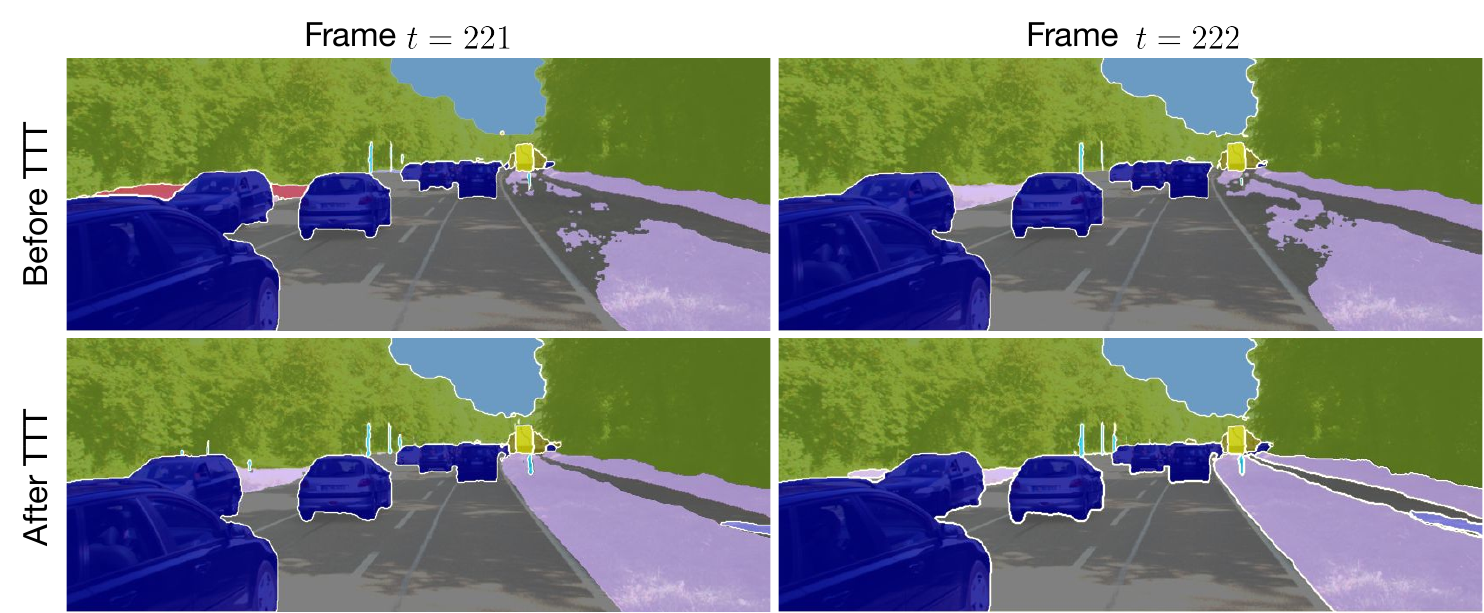}
    \caption{\footnotesize
    Semantic segmentation predictions for adjacent frames from a video in KITTI-STEP.
    \textbf{Top}:
    Results using a fixed model baseline without TTT. 
    Predictions are inconsistent between the two frames.
    The terrain on the right side of the road is incompletely segmented in both frames, and the terrain on the left is incorrectly classified as a wall on the first frame.
    \textbf{Bottom}: Results using \emph{Online TTT-MAE}, by the same model, on the same frames as top.
    Predictions are now consistent and correct.
    }
    \label{fig-kitti}
\end{figure*}
\begin{figure*}
    \centering
    \includegraphics[width=\textwidth]{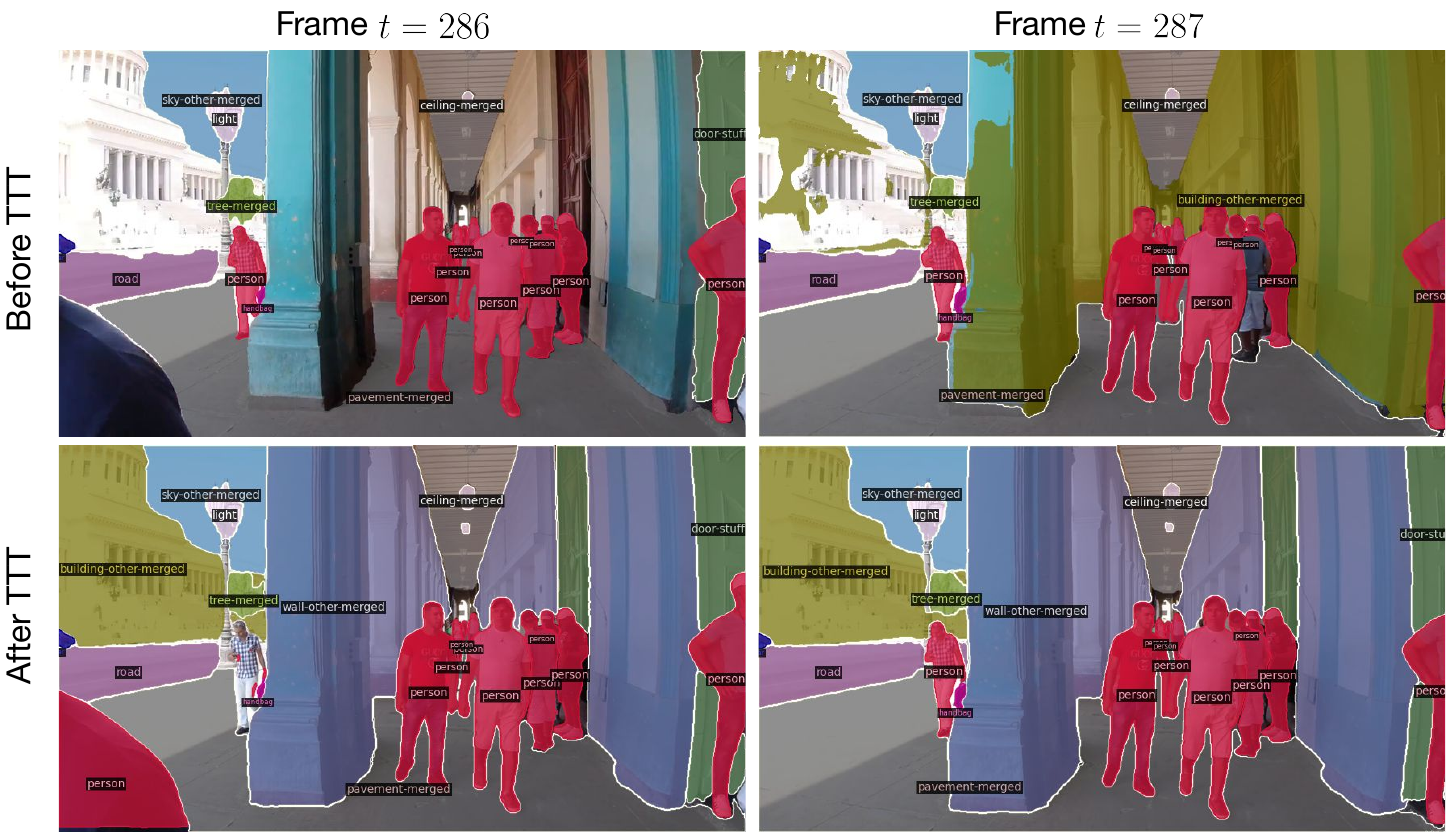}
    \caption{\footnotesize
    Panoptic segmentation predictions for adjacent frames from a video in our new COCO Videos dataset.
    \textbf{Top}:
    Results using a fixed model baseline without TTT. 
    Predictions are inconsistent between the two frames.
    \textbf{Bottom}: Results using \emph{Online TTT-MAE}, by the same model, on the same frames as top.
    Predictions are now consistent and correct.
    Please zoom in to see the instance labels.
    }
    \label{fig-coco}
\end{figure*}

\clearpage
\begin{figure*}
    \centering
    \includegraphics[width=0.75\textwidth]{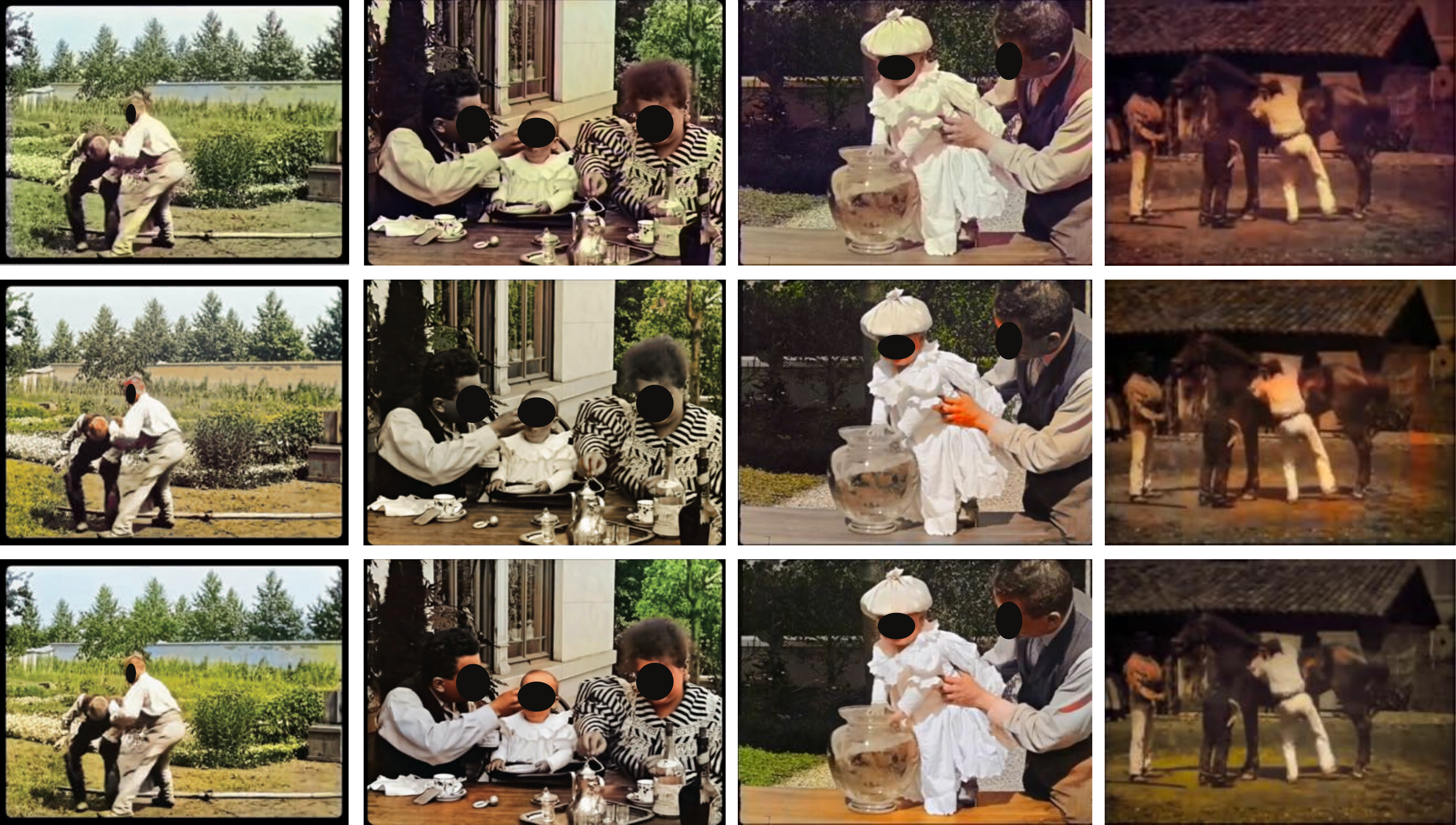}
    \caption{\footnotesize
    Samples results for video colorization on the Lumiere Brothers films.
    \textbf{Top}: Using Zhang et al.~\citep{zhang2016colorful}.
    \textbf{Middle}: Using our own baseline, Mask2Former with \emph{Main Task Only}, which is already comparable, if not superior to \cite{zhang2016colorful}.
    \textbf{Bottom}: After applying \emph{Online TTT-MAE} on top of the baseline.
    Our colors are more vibrant and consistent within regions.
    }
    \label{fig:colorization}
\end{figure*}

\section{Colorization Dataset - Lumi\`ere Brothers Films}
\label{app-color}
We provide results on the following 10 Lumiere Brothers films, all in the public domain:

\begin{enumerate}
    \item 
        Workers Leaving the Lumiere Factory (46 s)
    \item 
        The Gardener (49 s)
    \item 
        The Disembarkment of the Congress of Photographers in Lyon (48 s)
    \item 
        Horse Trick Riders (46 s)
    \item 
        Fishing for Goldfish (42 s)
    \item 
        Blacksmiths (49 s)
    \item 
        Baby's Meal (41 s)
    \item 
        Jumping Onto the Blanket (41 s)
    \item 
        Cordeliers Square in Lyon (44 s)
    \item 
        The Sea (38 s)
\end{enumerate}

\clearpage
\section{Proof of Theorem 1}
\label{app-proof}
We first prove the following lemma.

\textbf{Lemma.}
Let $f: \R^n \rightarrow \R$ be $\alpha$-strongly convex and continuously differentiable, and denote its optimal solution as $x^*$.
Let 
\begin{equation}
    \tilde{f}(x) = f(x) + v^Tx,
\end{equation}
and denote its optimal solution as $\tilde{x}^*$.
Then
\begin{equation}
    f(\tilde{x}^*) - f(x^*) \leq \frac{1}{2\alpha} \|v\|^2.
\end{equation}

\textbf{Proof of lemma.}
It is a well known fact in convex optimization \citep{bubeck2015convex} that for $f$ $\alpha$-strongly convex and continuously differentiable,
\begin{equation}
\label{strong-convexity}
    \alpha (f(x) - f(x^*)) \leq \frac{1}{2} \|\nabla f(x)\|^2,
\end{equation}
for all $x$.
Since $\tilde{x}^*$ is the optimal solution of $\tilde{f}$ and $\tilde{f}$ is also convex, we have $\nabla\tilde{f}(\tilde{x}^*) = 0$. But
\begin{equation}
    \nabla\tilde{f}(x) = \nabla f(x) + v,
\end{equation}
so we then have 
\begin{equation}
    \nabla f(\tilde{x}^*) = \nabla\tilde{f}(\tilde{x}^*) - v = -v.
\end{equation}
Make $x = \tilde{x}^*$ in Equation \ref{strong-convexity}, we finish the proof.

\textbf{Proof of theorem.}
By Assumptions 1 and 2, we have
\begin{equation}
    \|\nabla \ell_m^{~t}(\theta) - \nabla \ell_m^{~t-1}(\theta)\| \leq \beta\eta.
\end{equation}
\begin{align}
\frac{1}{k}
\sum_{t'=t-k+1}^t
\nabla\ell_s^{~t'} 
&= \frac{1}{k} \sum_{t'=t-k+1}^t \nabla\ell_m^{~t'} + 
\frac{1}{k} \sum_{t'=t-k+1}^t \delta_{t'} \\
&= \frac{1}{k} \sum_{t'=t-k+1}^t 
\left[\nabla\ell_m^{~t} 
+ \sum_{t''=t'}^{t-1} \left( \nabla\ell_m^{~t''} - \nabla\ell_m^{~t''+1} \right) \right] + 
\frac{1}{k} \sum_{t'=t-k+1}^t \delta_{t'} \\
&= \nabla\ell_m^{~t} + 
\frac{1}{k} 
\left[~
\sum_{t'=t-k+1}^t 
\sum_{t''=t'}^{t-1} \left( \nabla\ell_m^{~t''} - \nabla\ell_m^{~t''+1} \right) + 
\sum_{t'=t-k+1}^t \delta_{t'}
~\right]
\end{align}
To simplify notations, define
\begin{align}
    A &= \sum_{t'=t-k+1}^t 
\sum_{t''=t'}^{t-1} \left( \nabla\ell_m^{~t''} - \nabla\ell_m^{~t''+1} \right),\\
B &= \sum_{t'=t-k+1}^t \delta_{t'}.
\end{align}
So 
\begin{equation}
\frac{1}{k}
\sum_{t'=t-k+1}^t
\nabla\ell_s^{~t'} 
- \nabla\ell_m^{~t} = (A+B) / k.
\end{equation}
Because $\ell_m^{~t}$ is convex in $\theta$, we know that taking gradient steps with $\nabla\ell_m^{~t}$ would eventually reach the local optima of $\ell_m^{~t}$.
Because 
$\frac{1}{k}
\sum_{t'=t-k+1}^t
\nabla\ell_s^{~t'}$ differs from $\nabla\ell_m^{~t}$ by $(A+B) / k$, we know that taking gradient steps with the former reaches the local optima of 
$\ell_m^{~t} + (A+B)\theta / 2$.
Now we can invoke our lemma. To do so, we first calculate
\begin{align}
  \E \bigg\| \frac{1}{k}
\sum_{t'=t-k+1}^t
\nabla\ell_s^{~t'} 
- \nabla\ell_m^{~t} \bigg\|^2
 &= \frac{1}{k^2} \E \|A + B\|^2 \\
 &= \frac{1}{k^2} \left(\|A\|^2 + \E\|B\|^2 + \E A^TB\right) \\
 &\leq \frac{1}{k^2} \left(k^4\beta^2\eta^2 + k\sigma^2\right) \\
 &= k^2\beta^2\eta^2 + \frac{1}{k}\sigma^2.
\end{align}
Then by our lemma, we have
\begin{equation}
    \E\left[
    \ell_m(x_t, y_t; \tilde{\theta}) - \ell_m^*
    \right]
    \leq 
    \frac{1}{2\alpha}
    \E \bigg\| \frac{1}{k}
\sum_{t'=t-k+1}^t
\nabla\ell_s^{~t'} 
- \nabla\ell_m^{~t} \bigg\|^2 
\leq 
\frac{1}{2\alpha}
\left(
k^2\beta^2\eta^2 + \frac{1}{k}\sigma^2
\right).
\end{equation}
This finishes the proof.

\end{document}

%% file: notations.tex
\usepackage{mathtools}

\newcommand{\R}{\mathbb{R}}

\newcommand{\E}{\mathop{\mathbb{E}}}

\usepackage{mathtools}

\DeclareMathOperator*{\argmin}{arg\,min}

\usepackage{bm}